# Domain adaptation of a video Foundation model for Heart Failure Diagnosis from Echocardiography

Mahyar Bolhassani[1], Tijana Geroski[2], Narayana S. Singam[1], and Amir Amini, *IEEE Fellow*[1]
[1]Department of Electrical and Computer Engineering, University of Louisville, Louisville, KY
[2]Faculty of Engineering, University of Kragujevac, Kragujevac, Serbia

***Abstract*— The rapid advancement of foundation models has transformed computer vision, yet their adaptation to medical imaging, particularly purpose-built foundation models for spatio-temporal echocardiography videos, remains limited. In this work, we explore the feasibility of fine-tuning a pretrained foundation model for video (TimeSformer) for automated diagnosis of reduced ejection fraction heart failure (HFrEF) from apical four-chamber videos. The model is fine-tuned on the publicly available RVENet A4C dataset with a phase-aware clip generation strategy that captures complementary systolic and diastolic dynamics. We employ divided space–time attention, layer-wise learning rate decay, and multi-clip inference to enhance robustness under limited data. Experimental results show that full fine-tuning of the proposed model outperforms classifier-only tuning and a 3D CNN baseline. These findings highlight the potential of leveraging foundation video models for echocardiography-based heart failure diagnosis and motivates further domain adaptation studies in medical imaging.**

## I. Introduction

Heart failure is a major and growing public health concern that affects over 50 million people worldwide[1]. It is a complex clinical syndrome that can result from any structural or functional cardiac disorder that impairs the ability of the ventricle to fill with or eject blood. Heart failure with reduced ejection fraction (HFrEF) occurs when the left ventricular ejection fraction (LVEF) decreases to less than 40% [2]. Echocardiography is the primary imaging modality for evaluating cardiac structure and function due to low cost, accessibility, and safety [3]. However, manual interpretation remains time-consuming, and prone to inter/intra-observer variability, motivating the development of automated, and reproducible methods for detection of global and regional left ventricular systolic dysfunction - specifically HFrEF- classification [4]. Deep learning (DL) techniques have demonstrated strong potential in this domain, enabling automatic view classification, LV volume estimation, and systolic function assessment from echocardiographic data [5]. Conventional 2D parameters such as left ventricular ejection fraction (LVEF) remain gold standard for the diagnosis of HFrEF [6],[7][8], [21]. A fully automated echocardiogram interpretation in clinical practice considering multiple tasks including segmentation of cardiac chambers across 5 common views has been explored using 2D CNN pipeline in [9]. A video-based model has been utilized in [10], in which a 3D CNN model captures spatiotemporal patterns in videos. In [11], we proposed a 3D CNN pipeline for three-way heart disease classification from echocardiographic videos.

Recent advances in foundation models (FMs) and large vision models (LVMs), such as Vision Transformers (ViTs) [12] and video transformers like the TimeSformer [13], ViViT [14], Video Swin Transformer [15], and X-Clip [16] have rapidly reshaped the landscape of visual representation learning. These models, pretrained on massive and diverse datasets, capture hierarchical spatiotemporal relationships and can be efficiently fine-tuned for domain-specific tasks even with limited labeled data. However, within the medical imaging domain, such developments remain nascent—to the best of our knowledge, no foundation model has yet been pretrained on echocardiographic videos. Existing approaches, such as EchoCLIP [17], operate on 2D image-level representations rather than full spatiotemporal video, limiting its ability to model cardiac motion. Consequently, leveraging and adapting general-purpose foundation models for echocardiography remains challenging due to substantial modality gaps between natural videos and ultrasound data.

Nevertheless, exploring domain adaptation and advanced fine-tuning strategies presents a promising path forward. Fine-tuning can bridge the domain gap and enable effective transfer of general spatiotemporal knowledge into the medical imaging domain, where assembling large-scale video datasets for pretraining is often not feasible. In this work, we adapted a pretrained TimeSformer model originally trained on the Something-Something V2 (SSv2) dataset [18].When trained on the Something-Something V2 dataset, TimeSformer learns rich spatiotemporal representations encoding both appearance and motion cues. These representations can then be used for downstream tasks such as video captioning and perception of dynamic scenes. In our case we adapted the pretrained TimeSformer to the downstream task of binary classification of echocardiographic clips for distinguishing HFrEF from healthy subjects. However, future applications for this technique are numerous. Although this study primarily functions as a binary classification of HFrEF, there is a significant benefit in general-purpose video foundation models translating echocardiographic data with temporal sampling to assess specific cardiac pathologies. These include assessment of pericardial disease (i.e., pericardial tamponade versus constriction), left atrial and ventricular filling pressures, and infiltrative cardiomyopathies. Video foundation models can also be beneficial in the assessment of regional systolic dysfunction, something that would not be feasible with two-dimensional image processing. Lastly, this technique would be helpful in assessing and stratifying severity of ventricular desynchrony to further assist with planning for ventricular resynchronization therapies such as biventricular internal cardiac defibrillator implantation.

## II. Data

We used the public RVENet A4C dataset [19], which contains 3,583 2D A4C DICOM videos from 944 exams of 831 individuals between 2013 and 2021 on GE Vivid and Philips systems; from RVENet we used only the data from Healthy and HFrEF subjects. Fig. 1. Examples of a frame of an apical four-chamber view video of a healthy volunteer, the left one, and a subject with heart failure on the right [19]. In table I, the number of samples per class and train, validation, and test sets- are shown.

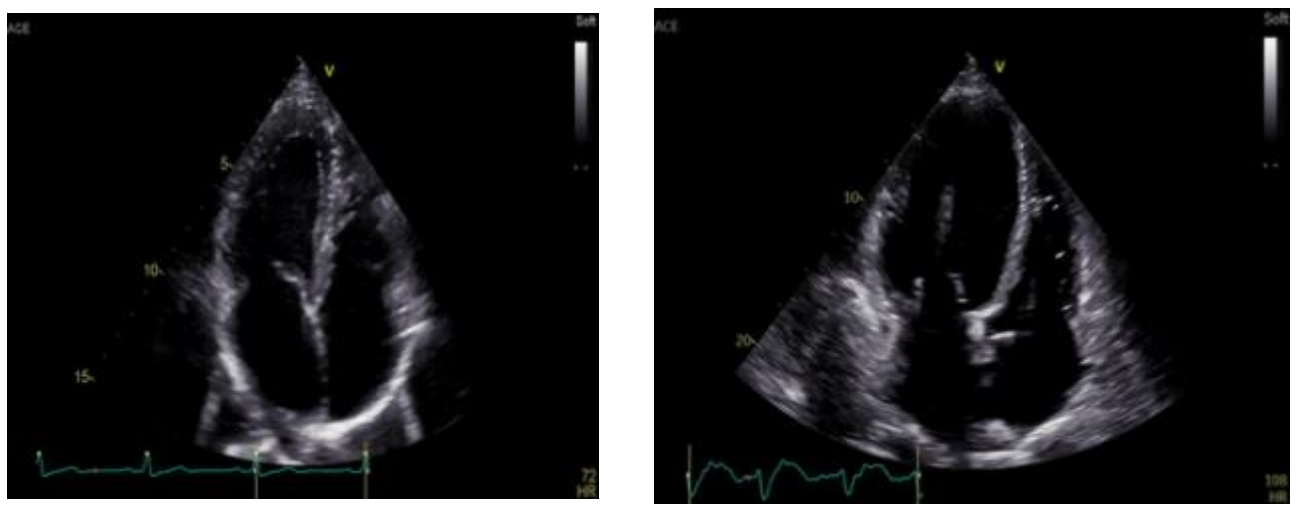

Fig. 1. sample frame of a healthy individual(left); sample frame of a HF volunteer (righ) [19]

TABLE I. Dataset

| Split | Healthy | HFrEF | Total |
|---|---|---|---|
| Train (80%) | 238 | 141 | 379 |
| Val (20%) | 59 | 35 | 94 |
| Test | 84 | 62 | 146 |
| Total | 381 | 238 | 619 |

### A. Preprocessing:

Videos were converted from YBR_FULL_422 to RGB. For temporal standardization, we estimate the single-cycle length from DICOM tags as $\mathbf{L} \approx \mathbf{FPS} \times \mathbf{60/HR}$ , partitioning each video into complete cycles. For spatial standardization, we applied motion-based cropping: computed a motion mask from frame-to-frame differences, extracted the minimal bounding box around the dynamic cardiac region, expanding it slightly to preserve context. These steps serve to reduce spurious background features outside the scanned area. Fig. 2. Shows preprocessed samples with their corresponding motion-based masks.

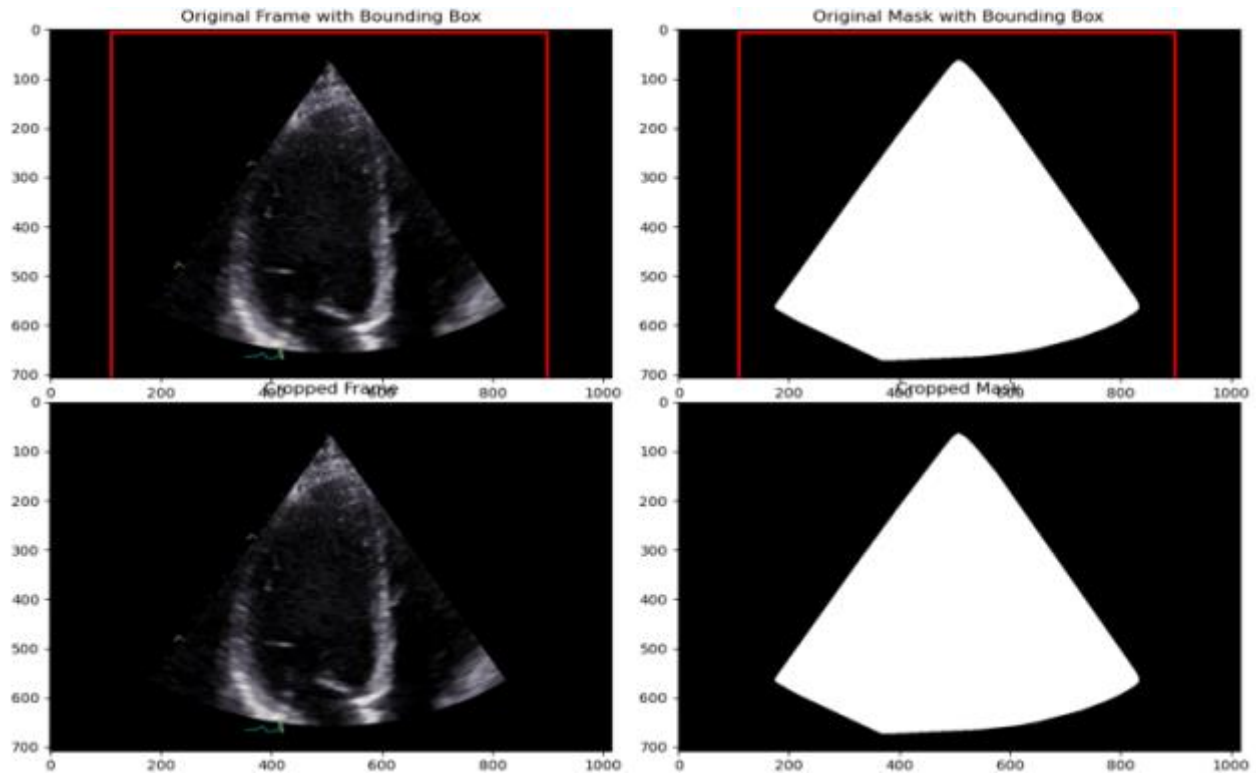

Fig. 2. preprocessed samples with corresponding masks

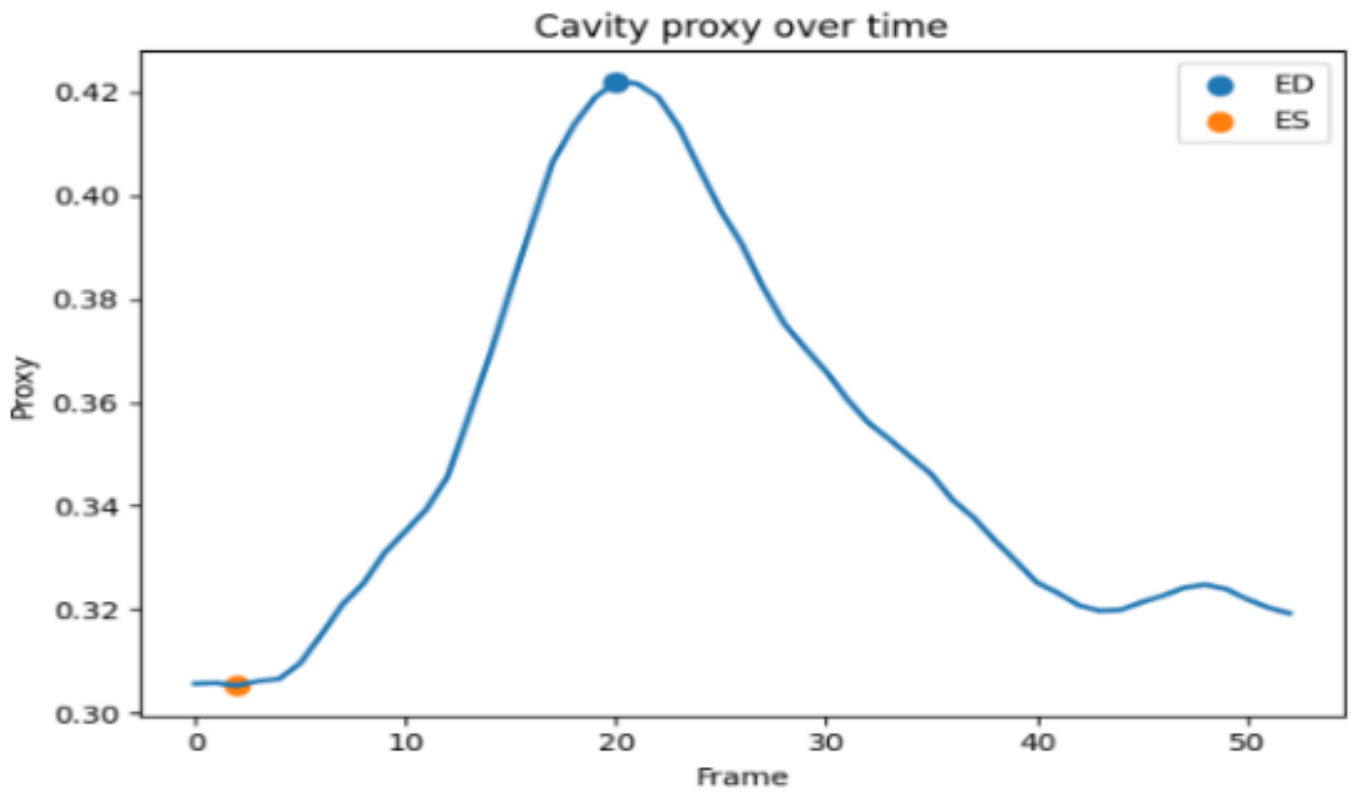

Fig. 3. ED/ES estimation by cavity proxy calculation over time

*B. Cardiac phase localization: extracting end-diastole (ED) and end-systole (ES)*

In order to extract shorter yet cycle aware clips out of each video while maintaining compatibility with the pretrained model, we need to automatically identify the ED/ES frames. We localize ED/ES on a per-cycle basis using an LV-size proxy computed from the region of interest (ROI). Within a cycle, we compute a per-frame scalar signal s(t) that which tracks LV cavity size. Since LV masks are not available, we use a segmentation-free proxy defined as motion energy inside the ROI: $s(t) = \sum_{x \in ROI} \|\nabla I_t(x)\|_2$; Where $I_t$ is the image intensity at pixel x in frame t and $\nabla I_t(x)$ indicates the spatial gradient at x in frame t.

We take ED to be the time t that maximizes $s(t)$ (largest LV chamber size in the video) and ES at the time that minimizes s(t) (smallest LV chamber), ensuring ED precedes ES in the beat. We suppressed spurious peaks, short, low-prominence local extrema in s(t) caused by noise/artifacts rather than true cardiac phases, using a 1.5 SD prominence threshold[20]. Fig. 3 shows the plot of cavity size proxy s(t) over time for a sample video.

*C. Multi-clip selection*

TimeSformer [13] models pre-trained on SSv2[18] provided by HuggingFace, require fixed-length clips (N ∈ $\{8, 16\}$ ). We generated several complementary clips per cardiac cycle and aggregated predictions across them.

1. Phase-anchored windows (ED/ES centric): For each cycle, we extracted 2 clips of length N centered at the ED and ES indices. If a clip runs past the start or end of the cycle, we use linear temporal interpolation to keep the number of frames to be exactly N.
2. Phase-spanning resample (uniform): We uniformly resampled the entire cycle to N frames, preserving beat-scale dynamics regardless of native frame count. This clip provides global context across isovolumetric phases and valve events.
3. Multi-frame sampling: we adapt EchoCLIP[17] strategy of sampling to TimeSformer by sampling one random temporal clip per epoch for training and aggregating predictions across different sampled clips at reference.

## III. METHOD

Among the available video-based transformer architectures, TimeSformer was chosen for this study due to its architectural simplicity and the readily accessible pretrained weights, including officially released checkpoints via the Hugging Face Transformers library.

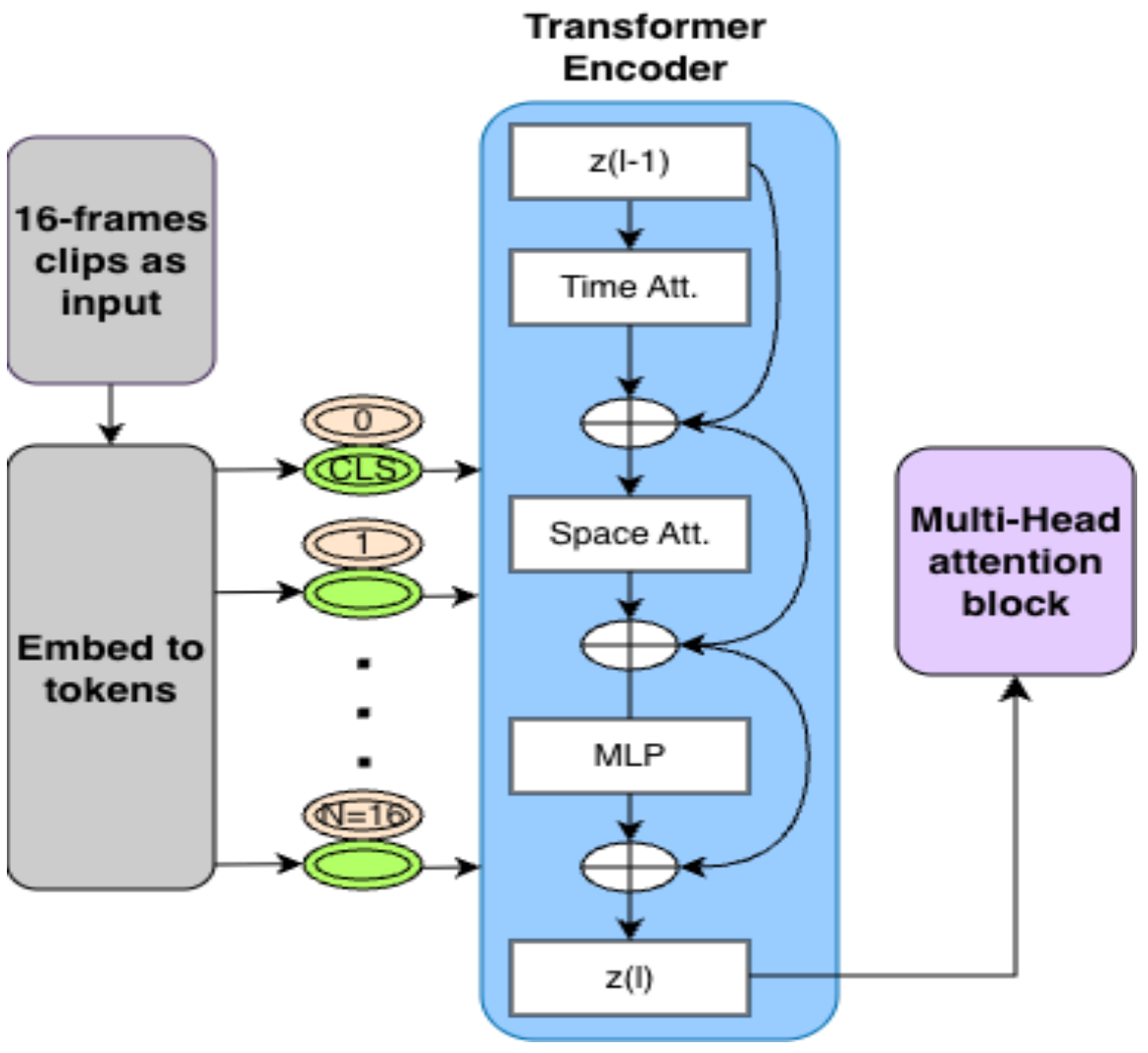

Fig. 4. Phase-aware TimeSformer [12] encoder with divided space–time attention. Distinct T+S attention efficiently models motion and anatomy separately.

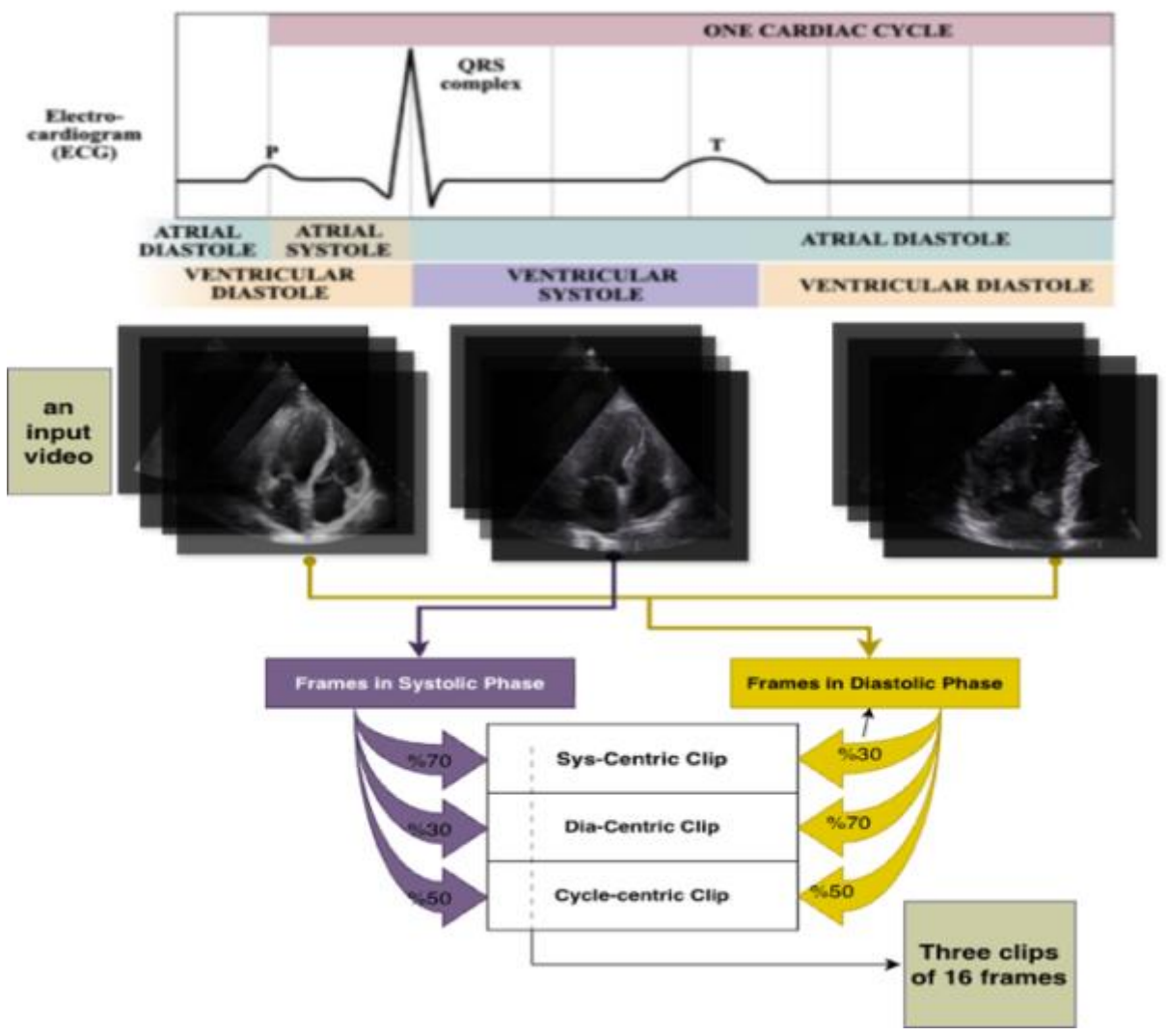

Fig. 5. Clip construction across cardiac phases; Generation of systolic, diastolic, and cycle-centric 16-frame clips from a single cardiac cycle

As the first fully transformer-based framework to successfully model video sequences through a divided space–time attention mechanism—comprising distinct temporal and spatial attention stages (Time Att. and Space Att.), as illustrated in Fig. 4, TimeSformer offers a transparent and modular representation of spatiotemporal dependencies, making it particularly well-suited for medical video analysis.

## A. *Clip Generation and Phase selection:*

Given echocardiography cine apical views $\chi = \Re^{T\times C\times H\times W}$, the objective is binary classification $y \in \{0, 1\}$ where y = 1 indicates HFrEF (LVEF $\leq$ 40%). A transformer-based video architecture with spatiotemporal self-attention was employed to learn discriminative disease representations.

As illustrated in Fig.5, each input video is first segmented into individual cardiac cycles. As described in section II.B, frames corresponding to systolic and diastolic periods are then separated to capture phase-specific dynamics. Temporal ratios (e.g., 70% systolic/30% diastolic for systole-centric clips and 70% diastolic/30% systolic for diastole-centric clips) are adjusted to ensure coverage of both contraction and relaxation phases while keeping a fixed temporal length T=16. Each frame was resized to (224,224), and normalized to [0,1]. The resulting 16-frame clips provide consistent temporal context for training and evaluation.

## B. *Architecture:*

The adopted backbone is the TimeSformer model [14] using divided space-time attention (T+S) variant, illustrated in Fig. 3. Clips are tokenized into non-overlapping 16×16 patches per frame. Each patch is linearly projected to d=768 embedding; a learnable token summarizes the sequence.

Learnable spatial and temporal positional embeddings are added. Within each transformer block, temporal attention is first applied across frames followed by spatial attention within each frame, enabling efficient modeling of both motion and anatomy. Model weights were initialized from a large-scale video-pretrained checkpoints (SSv2[18]) and the classification head was replaced with two-class layer. All transformer layers were unfrozen during fine-tuning to adapt spatio-temporal representations to the echocardiograph domain. To address moderate class imbalance and emphasize difficult-to-classify samples, a binary focal loss was adopted which reduces the influence of easy samples while preserving balanced gradient contribution between HFrEF and Healthy samples.

# IV. IMPLEMENTATION DETAILS

Training was performed using the AdamW optimizer with a base learning rate of $1\times10^{-4}$ for pretrained transformer layers and $1\times10^{-3}$ for the classification head. A layer-wise learning rate decay of 0.75 was applied, assigning smaller learning rates to shallower layers to preserve general representations learned during pretraining while allowing deeper layers to adapt more aggressively to domain-specific cardiac features.

TABLE II. PERFORMANCE COMPARISON OF FINE-TUNING TIMESFORMER

| Model | ROC-AUC | F1 | Precision | Accuracy | epochs |
|---|---|---|---|---|---|
| ***Phase-aware TimeSformer (Full FT)*** | ***0.893*** | ***0.783*** | ***0.789*** | ***0.763*** | ***11*** |
| ***Phase-aware TimeSformer (NFT)*** | ***0.783*** | ***0.702*** | ***0.686*** | ***0.711*** | ***12*** |
| ***ResNet18*** | ***0.849*** | ***0.732*** | ***0.764*** | ***0.732*** | ***26*** |

Each A4C video was preprocessed by cropping the ultrasound sector, and normalizing pixel intensities to the range [0,1]. From each video, multiple 16-frame clips were generated with random starting points and uniform stride; where possible, frame overlap between clips was avoided to increase data diversity and reduce redundancy.

Training used a batch size of 8 and was conducted for 50 epochs, with early stopping (patience = 3) based on the validation F1-score to prevent overfitting. During inference, as shown in Fig. 6, we employed a multi-clip ensemble strategy to improve prediction robustness and mitigate variability due to temporal sampling. For each echocardiographic study, three non-overlapping clips were uniformly sampled to capture distinct temporal segments of the cardiac cycle, ensuring diversity in motion representation across clips. Each clip was independently processed by the model to produce a class probability. Results

We conducted a comparative evaluation between three models: (1) Phase-aware TimeSformer with full fine-tuning (Fully FT), (2) Phase-aware TimeSformer with classifier-head-only fine-tuning (NFT) where backbone remains frozen, and (3) a 3D CNN baseline - 3D ResNet18, a widely adopted model in medical imaging research. All models were trained under identical hyperparameter settings to ensure a fair comparison across architectures and adaptation strategies.

In addition to the performance comparison summarized in Table II, Fig. 6 presents the ROC curves for the selected TimeSformer fine-tuning configurations and the 3D ResNet18 baseline. Each curve plots the Receiver's Operating Characteristic, displaying the model's trade-off between True Positive Rate and False Positive Rate at various decision thresholds, with the area under each curve (AUC) reported in parentheses.

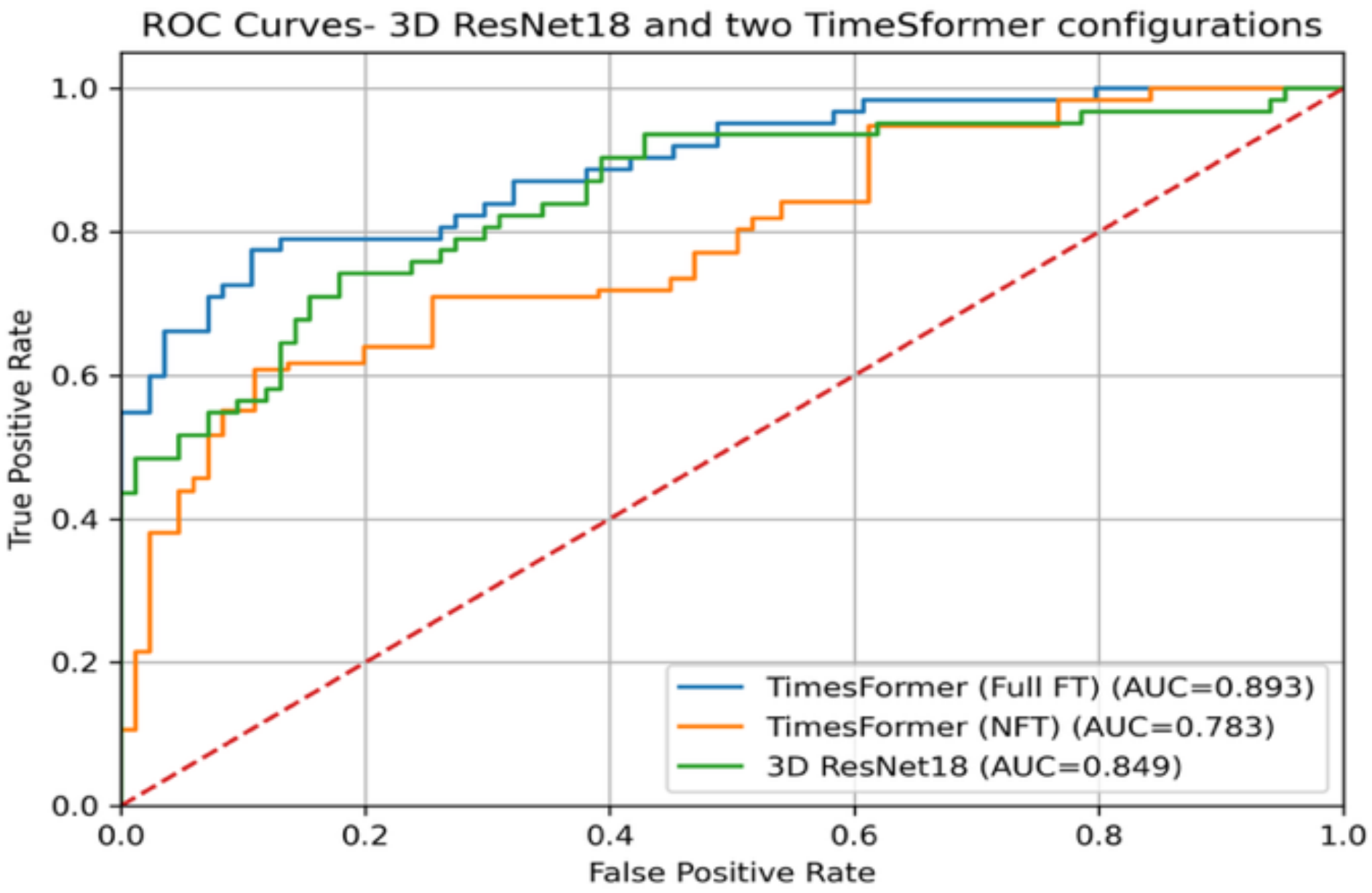

Fig. 6. Receiver Operating Characteristic (ROC) curves, blue for TimeSformer with full fine-tuning, green for 3D ResNet18, and orange for TimeSformer with classifier head only fine-tuning.

Fig.6. visually confirms the performance trends highlighted in Tables II, where phase-aware TimeSformer with full fine-tuning pushes the curve closer to the top-left corner (indicating higher True Positive rate and lower False Positive Rates for the classifier).